\documentclass[conference]{IEEEtran}

\usepackage[T1]{fontenc}
\usepackage{times}
\usepackage{graphicx}
\usepackage{amsmath,amssymb}
\usepackage{booktabs}
\usepackage{stfloats}
\usepackage{balance}
\usepackage[dvipsnames]{xcolor}
\usepackage[breaklinks,colorlinks,allcolors=blue]{hyperref}
\usepackage[capitalize]{cleveref}

\begin{document}

\title{Fusion Complexity Is Not Fusion Quality:\\Backbone, Cross View Fusion and Metadata as Separate Factors in Pasture Biomass Regression}

\author{
  \IEEEauthorblockN{Mridankan Mandal}
  \IEEEauthorblockA{Department of Information Technology\\
  Indian Institute of Information Technology, Allahabad\\
  Prayagraj, India\\
  mridankanmandal2006@gmail.com}
}

\maketitle

\begin{abstract}
Estimating pasture biomass from field images can support grazing decisions without repeated destructive sampling.
The task remains difficult because available datasets are small, biomass components are imbalanced and a single photograph contains two spatially distinct views of the sampled quadrat.
This work evaluates vision foundation models for five target pasture biomass regression on the 357 image CSIRO dataset.
Across 25 model configurations, the analysis covers convolutional and self supervised transformer backbones, seven cross view fusion configurations, models with and without 23 metadata features, classical regressors and frozen foundation features.
All comparisons use one locked five fold split and raw scale weighted $R^2$.
DINOv3 Large with cross view gated attention gives the strongest verified single model result, $0.8338\pm0.0503$ across folds.
Under a matched gated depthwise convolution head, replacing DINOv2 Large with DINOv3 Large improves $R^2$ by 0.0372.
Fusion results are not ordered by module complexity.
Cross view gated attention is best, while a full Mamba fusion block reaches only 0.7930.
Across four matched DINOv3 pathways, metadata narrows the score range from 0.0157 to 0.0013 while changing the mean by only $\mathbf{+0.0034}$.
It improves three weaker pathways but lowers the strongest attention model by 0.0061.
Classical metadata models are strong, frozen DINOv3 features are linearly useful and Dry Dead biomass remains the hardest target.
These results show that backbone choice, cross view interaction and metadata must be tested as separate factors on small agricultural datasets.

\end{abstract}

\section{Introduction}
\label{sec:intro}

Pasture biomass is a direct measure of feed availability and an important input to grazing management.
The reference procedure cuts, sorts, dries and weighs vegetation within a quadrat.
It gives physically meaningful measurements but it is destructive and costly to repeat over large areas.
Image based regression offers a practical alternative if the model can separate green vegetation, dead material and clover under changes in species, season and field conditions.

Agricultural image regression has two connected difficulties.
First, labelled field data are expensive.
The 357 image CSIRO dataset used here is small, yet each image is large and visually heterogeneous.
Second, biomass is not a single visual quantity.
Plant colour, texture, occlusion, height, moisture and botanical composition jointly affect the targets.
Simple vegetation indices can describe part of this variation, while learned representations can capture richer structure~\cite{lu2006remote,bannari1995review,murphy2024deep}.

Large self supervised vision models provide a useful starting point for this setting.
DINO, DINOv2 and DINOv3 learn visual features without task labels and transfer across image domains~\cite{caron2021emerging,oquab2024dinov2,simeoni2025dinov3}.
Their use does not remove the need for task design.
The two halves of a pasture photograph still need to be combined, predictions must obey biomass composition and metadata may add information that is not visible in the image.
These choices can matter as much as the regression head when the training set is small.

This paper studies three questions.
Which backbone transfers best to pasture biomass regression?
How should two image views interact?
When does tabular metadata help?
The questions are evaluated under a locked five fold protocol with one common metric and the same image split.
The comparison includes EfficientNet, DINOv2, DINOv3 and VMamba models, identity, convolutional, attention and state space fusion and eight classical or frozen feature baselines.

The main findings are:

\begin{itemize}
    \item DINOv3 Large is the strongest tested backbone. With the same two block gated depthwise fusion head, it improves mean fold $R^2$ from 0.7902 to 0.8274 over DINOv2 Large.
    \item Fusion quality does not follow a simple complexity order. Cross view gated attention reaches 0.8338 but full Mamba fusion reaches 0.7930 and falls below identity addition at 0.8181.
    \item Metadata has an architecture dependent effect. Across four matched DINOv3 pathways, it reduces the score range by 91.7 percent, from 0.0157 to 0.0013, while changing the mean by only +0.0034.
    \item Metadata alone supports competitive classical prediction. Frozen DINOv3 features also give a strong linear baseline but adapting the image encoder and learning cross view interactions retain a clear advantage.
    \item Dry Dead is the hardest biomass component and fold 4 is the weakest fold for every selected high performing model.
\end{itemize}

Our goal is practical.
We compare backbone transfer, cross view fusion and metadata under the same split and metric.
We show where each factor helps and where it does not.
We do not claim that one fusion rule is best in every setting.

\section{Related Work}
\label{sec:related}

\textbf{Agricultural image regression.}
Remote sensing based biomass estimation has long combined spectral response, vegetation structure and statistical regression~\cite{lu2006remote}.
Vegetation indices summarize visible or multi-spectral signals but their relation to biomass can saturate and change across species and environments~\cite{bannari1995review}.
Combining plant height with visible or near infrared indices can improve crop biomass estimation over spectral indices alone~\cite{bendig2015combining}.
Modern agricultural vision increasingly uses deep learning to learn task specific features~\cite{kamilaris2018deep,murphy2024deep}.
Deep Plant Phenomics showed plant counting, classification and regression in a common software framework~\cite{ubbens2017deep}.
Examples include counting sorghum heads from aerial images~\cite{guo2018aerial} and estimating grass and clover composition from field photographs~\cite{skovsen2019grassclover}.
The CSIRO Pasture Biomass dataset extends this direction with laboratory measured component biomass, field metadata and data collected across 19 Australian locations~\cite{liao2025pasture}.

\textbf{Vision foundation models.}
Vision foundation models build on transformer tokenization and large scale pretraining.
ViT showed that image patches can be processed directly by a transformer, CLIP learned transferable image representations from image text pairs and masked autoencoders established scalable reconstruction based pretraining~\cite{dosovitskiy2021image,radford2021learning,he2022masked}.
SimCLR showed that contrastive objectives can learn transferable visual representations without class labels~\cite{chen2020simclr}.
DeiT showed that vision transformers can be trained more efficiently with restricted labelled data and transformer specific distillation~\cite{touvron2021deit}.
DINO showed that self distillation can produce transformer features with useful semantic structure without class labels~\cite{caron2021emerging}.
DINOv2 scaled this approach to a curated image corpus and produced general purpose visual features~\cite{oquab2024dinov2}.
DINOv3 further increased data and model scale while addressing feature degradation during long training~\cite{simeoni2025dinov3}.
These encoders are relevant to agricultural tasks because broad pretraining can reduce dependence on scarce field labels.
The present work compares supervised convolutional features, frozen foundation features, supervised probes and full model adaptation under one split.

\textbf{Cross view fusion.}
Multiple views can be combined by addition, concatenation, local convolution, cross attention, or sequence models.
Multi-modal learning separates representation, alignment and fusion and notes that sources may differ in predictive value and noise~\cite{baltrusaitis2019multimodal}.
This distinction matters here because the two image views share one sensor, while metadata forms a separate modality.
CrossViT exchanges information between token streams through attention~\cite{chen2021crossvit}, while multi-modal transformers use cross modal attention to align separate sequences~\cite{tsai2019multimodal}.
Gated convolution provides a local, parallel alternative~\cite{dauphin2017gated}.
Selective state space models provide content dependent sequence processing with linear scaling in sequence length~\cite{gu2023mamba}.
Their vision adaptations include VMamba and bidirectional Vision Mamba~\cite{liu2024vmamba,zhu2024visionmamba}.
This work treats these mechanisms as alternatives instead of assuming that greater sequence modelling capacity will improve a small regression task.

\section{Method}
\label{sec:method}

\subsection{Problem Formulation}
\label{sec:problem}

Each sample contains one wide pasture image and five biomass targets measured in grams: Dry Green, Dry Dead, Dry Clover, Green Dry Matter and Dry Total.
The last two targets are compositional:
\begin{equation}
\widehat{y}_{\mathrm{GDM}}=\widehat{y}_{\mathrm{Green}}+\widehat{y}_{\mathrm{Clover}},
\end{equation}
\begin{equation}
\widehat{y}_{\mathrm{Total}}=\widehat{y}_{\mathrm{GDM}}+\widehat{y}_{\mathrm{Dead}}.
\end{equation}
Models predict the three primitive components and form the two aggregate outputs by summation.
Softplus output activations keep primitive predictions non-negative.

The primary score is weighted coefficient of determination on raw biomass values:
\begin{equation}
R^2_w=\sum_{k=1}^{5}w_k\left(1-\frac{\sum_i(y_{ik}-\widehat{y}_{ik})^2}{\sum_i(y_{ik}-\overline{y}_k)^2}\right),
\label{eq:metric}
\end{equation}
where $w=[0.1,0.1,0.1,0.2,0.5]$.
Dry Total receives half of the score, Green Dry Matter receives one fifth and each primitive component receives one tenth.
This audited raw scale score differs from the log transformed score described for the public challenge~\cite{liao2025pasture}.
All numbers in this paper use \Cref{eq:metric}, so they should not be compared directly with log scale challenge scores.

\subsection{Dual View Regression Pipeline}
\label{sec:pipeline}

\begin{figure*}[!t]
\centering
\includegraphics[width=0.98\textwidth]{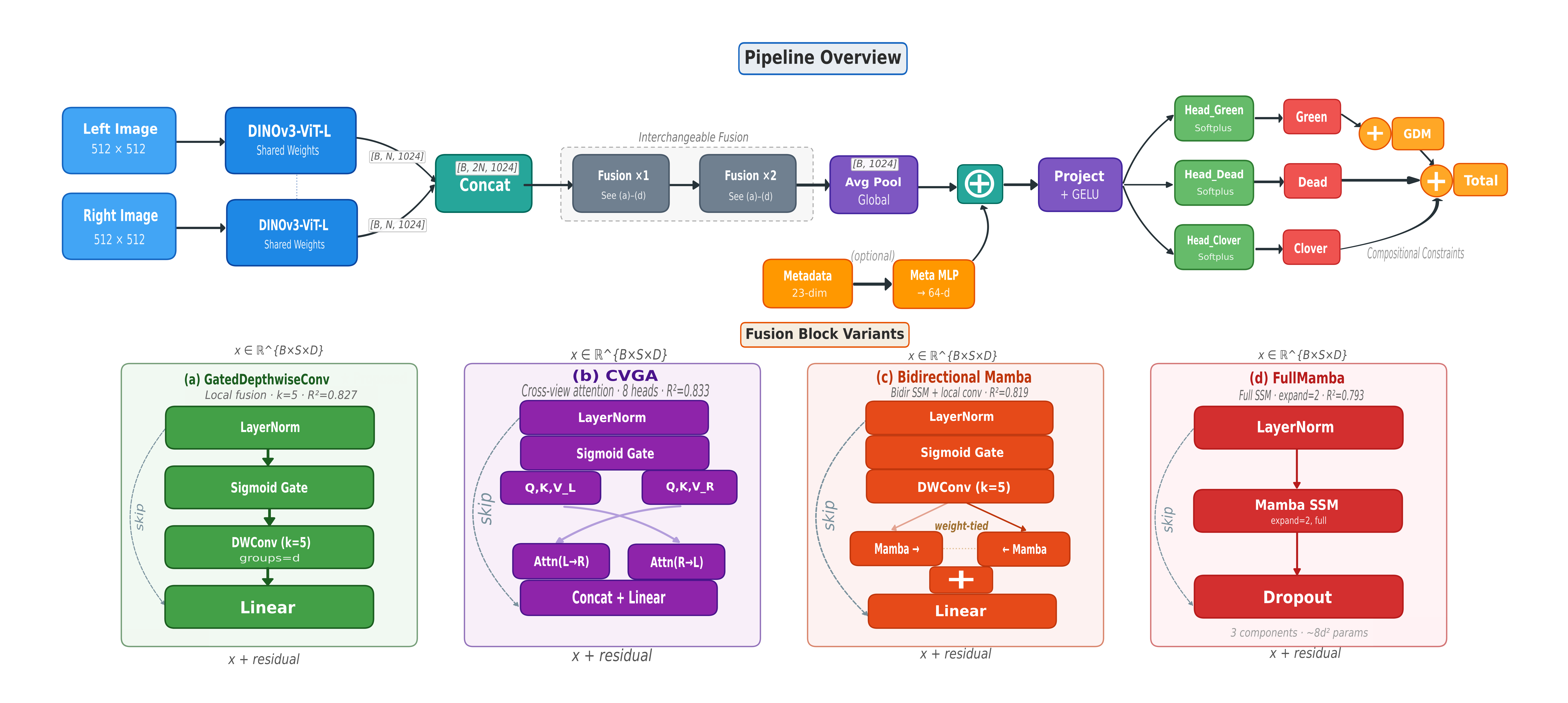}
\caption{Dual view biomass regression pipeline and the tested fusion modules. The same backbone encodes both image halves. Fusion is followed by pooling and three non-negative primitive heads. Green Dry Matter and Dry Total are obtained through fixed sums.}
\label{fig:architecture}
\end{figure*}

The wide image is split into left and right views.
Both views pass through one weight tied visual backbone.
The resulting patch token sequences are joined by a fusion module, average pooled and mapped to the three primitive targets.
For DINOv3 Large at $512\times512$ resolution, each view gives a $32\times32$ patch grid with 1024 dimensional tokens.
The shared encoder preserves view symmetry and avoids doubling the backbone parameters.
The complete data flow and fusion variants are shown in \Cref{fig:architecture}.

\subsection{Backbones and Fusion Modules}
\label{sec:fusion}

The backbone comparison includes EfficientNet B3~\cite{tan2019efficientnet}, DINOv2 Base, Large and Giant~\cite{oquab2024dinov2}, DINOv3 Large~\cite{simeoni2025dinov3} and VMamba Base~\cite{liu2024vmamba}.
Supervised probes use cached frozen embeddings.
Fusion experiments use the DINOv3 Large token grid unless stated otherwise.

Identity addition is the simplest fusion rule.
It adds aligned left and right tokens before pooling.
Gated depthwise convolution applies layer normalization, a learned sigmoid gate, a depthwise one dimensional convolution with kernel size 5, a linear projection, dropout and a residual connection.
One, two and four block variants test depth without changing the backbone.

Cross view gated attention separates the joint sequence into left and right streams.
Eight head cross attention is applied in both directions.
A learned sigmoid gate controls each cross view update before the residual addition.
Bidirectional Mamba processes the sequence in forward and reverse order with shared weights, while full Mamba uses a standard selective state space block~\cite{gu2023mamba,zhu2024visionmamba}.
These alternatives cover no learned fusion, local learned fusion, global attention and global state space fusion.
Their implemented structure and parameter counts are summarized in \Cref{tab:fusion_design}.
Two block capacity ranges from zero parameters for identity addition to 17.55 million parameters for bidirectional Mamba.
This range allows fusion behavior to be compared with both interaction scope and learned capacity stated explicitly.

\begin{table}[!t]
\centering
\caption{Implemented fusion modules at token width 1024. Counts exclude the backbone and prediction heads.}
\label{tab:fusion_design}
\resizebox{\columnwidth}{!}{%
\begin{tabular}{lccc}
\toprule
\textbf{Fusion} & \textbf{Token interaction} & \textbf{Params/block} & \textbf{Params/2 blocks} \\
\midrule
Identity addition & Aligned addition & 0 & 0 \\
GDWC & Local, kernel 5 & 2.11M & 4.21M \\
CVGA & Global cross attention & 5.25M & 10.50M \\
BidirMamba & Local and global scan & 8.77M & 17.55M \\
FullMamba & Global scan & 6.67M & 13.34M \\
\bottomrule
\end{tabular}%
}
\end{table}

After adaptive average pooling, each primitive head uses a two layer multi-layer perceptron with GELU, dropout 0.2 and Softplus output.
Models with metadata encode 23 standardized variables through a small multi-layer perceptron, concatenate the result with the pooled image representation and project the joint vector before the prediction heads.

\subsection{Classical and Frozen Feature Baselines}
\label{sec:classical}

Eight additional baselines measure how much information is available without adapting the image encoder.
Metadata features are evaluated with multi-output Ridge regression and independent XGBoost models.
XGBoost is a regularized tree boosting system designed for efficient learning on structured inputs~\cite{chen2016xgboost}.
Tree models remain strong controls for medium sized tabular datasets~\cite{grinsztajn2022tabular}.
Handcrafted image descriptors cover colour statistics, visible vegetation indices, texture and spatial variation from both views.
They are tested with Ridge and XGBoost, then combined with metadata for a multi-modal XGBoost baseline.
Frozen DINOv2 Base and DINOv3 Large embeddings are evaluated with Ridge regression.
Their inner search selects no PCA, 64 components, or 128 components separately for each outer fold.
The DINOv3 probe selects 128 components in two folds and no PCA in three folds.
The final baseline fixes DINOv3 features to 32 principal components before XGBoost.
Hyperparameters are selected only from each outer training partition through an inner three fold split.

\section{Experimental Setup}
\label{sec:setup}

\subsection{Dataset and Split}

The experiments use the 357 training images released through the CSIRO Image2Biomass challenge~\cite{liao2025pasture}.
This split provides all five laboratory biomass targets and the training metadata used in our experiments.
We evaluate every configuration on the same five folds so backbone, fusion and metadata comparisons use identical images and targets.
Each sample contains a top view of a $70\,\mathrm{cm}\times30\,\mathrm{cm}$ quadrat and 23 encoded metadata features derived from date, state, species, height and vegetation measurements.

The target statistics in \Cref{tab:target_stats} show strong imbalance across components.
Dry Clover has 37.8\% zero values and the largest skewness at 2.84.
Dry Total has no zero values and receives half of the evaluation weight.
Dry Green and Green Dry Matter have Pearson correlation 0.884, while Green Dry Matter and Dry Total have correlation 0.896, partly because the aggregate targets are defined by sums.
These properties make a single overall score sensitive to both target composition and the high biomass tail.

\begin{table}[!t]
\centering
\caption{Biomass target statistics for the 357 image CSIRO dataset. Values are in grams except skewness and zero rate.}
\label{tab:target_stats}
\resizebox{\columnwidth}{!}{%
\begin{tabular}{lrrrrrrr}
\toprule
\textbf{Target} & \textbf{Mean} & \textbf{SD} & \textbf{Min} & \textbf{Median} & \textbf{Max} & \textbf{Skew} & \textbf{Zero \%} \\
\midrule
Dry Green & 26.6 & 25.4 & 0.0 & 20.8 & 158.0 & 1.75 & 5.0 \\
Dry Dead & 12.0 & 12.4 & 0.0 & 8.0 & 83.8 & 1.76 & 11.2 \\
Dry Clover & 6.7 & 12.1 & 0.0 & 1.4 & 71.8 & 2.84 & 37.8 \\
Green Dry Matter & 33.3 & 24.9 & 1.0 & 27.1 & 158.0 & 1.56 & 0.0 \\
Dry Total & 45.3 & 28.0 & 1.0 & 40.3 & 185.7 & 1.43 & 0.0 \\
\bottomrule
\end{tabular}%
}
\end{table}

One five fold split is locked for all experiments.
Stratified group cross validation uses Dry Total quintiles for stratification and image identifiers for grouping.
The split seed is 17 and fold sizes are 72, 72, 71, 71 and 71.
Transform fitting, feature standardization, principal component analysis and classical model selection use training data only.

\subsection{Training and Evaluation}

Images are resized to the resolution required by each backbone, normalized with ImageNet statistics and augmented with standard geometric and colour transforms.
The two views are transformed separately.
Deep models use AdamW~\cite{loshchilov2019decoupled} with backbone learning rate $10^{-5}$, head learning rate $5\times10^{-4}$, weight decay $10^{-2}$, five warmup epochs and cosine decay.
Training lasts at most 50 epochs.
The loss is Smooth L1 with transition parameter 5, applied to raw gram targets~\cite{huber1964robust}.
Gradient norm is clipped at 1.0.
Mixed precision reduces training memory and computation cost~\cite{micikevicius2018mixed}.
Gradient checkpointing is also used to fit training on one NVIDIA GeForce RTX 4060 Laptop GPU.

The held out outer fold is also used for checkpoint selection.
This reuse can bias performance upward because model selection and assessment use the same held out data~\cite{varma2006bias}.
The reported folds are cross validation estimates and not untouched test results.
All reported deep configurations use one master seed, 17, so fold standard deviation measures variation across data partitions, not training seed uncertainty.
Fold scores are also dependent because their training partitions overlap, so this standard deviation is descriptive instead of a complete variance estimate~\cite{bengio2004variance}.
No test time augmentation is used by the common evaluator.
Mean fold raw $R^2_w$ is the primary comparison and pooled out of fold $R^2_w$ is reported for target analysis.

Metadata uses a different evaluation setup.
The public challenge withholds metadata for validation and test images~\cite{liao2025pasture}.
Our metadata enabled results provide metadata to each held out fold and form an analytical cross validation setting.
They measure conditional value when metadata exists at evaluation time, not performance under the challenge submission contract.

\section{Results}
\label{sec:results}

\subsection{Backbone and Baseline Comparison}

\begin{figure}[!t]
\centering
\includegraphics[width=\columnwidth]{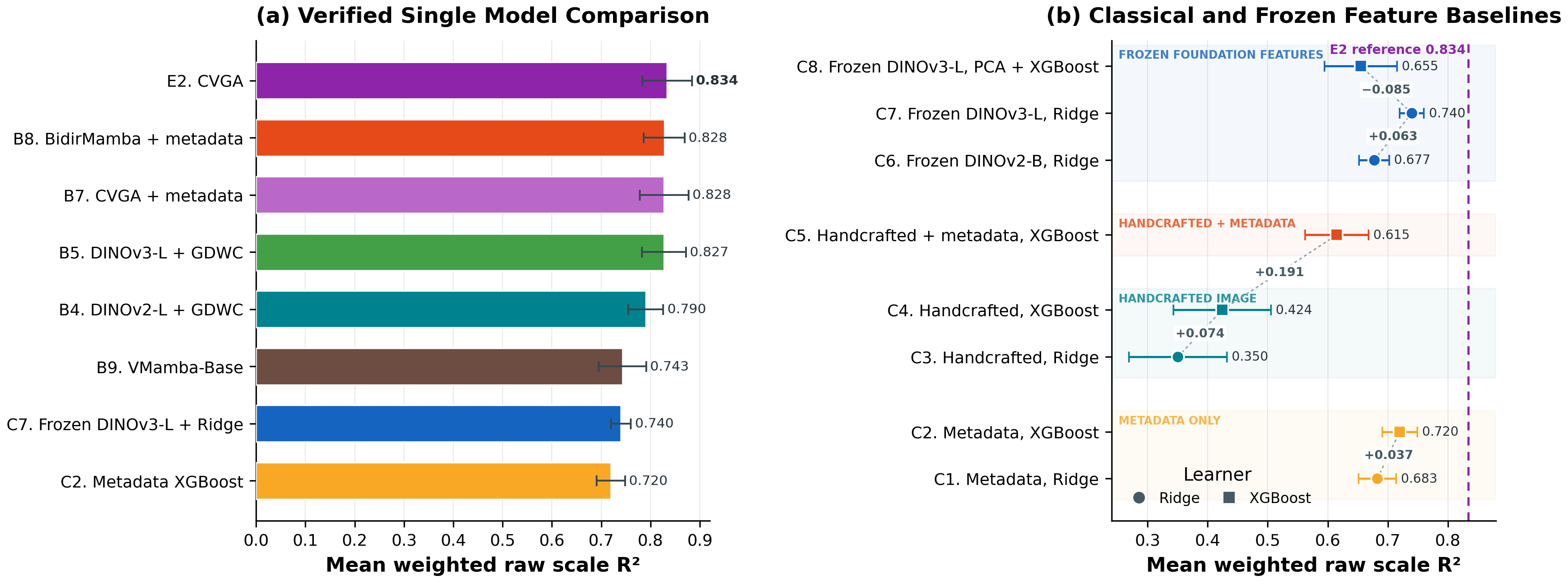}
\caption{Model comparison using mean five fold raw $R^2_w$. Panel (a) shows representative single models. Panel (b) shows classical and frozen feature baselines. Error bars are outer fold standard deviations. The dashed line marks E2 at 0.8338.}
\label{fig:overview}
\end{figure}

The model ranking in \Cref{fig:overview}a is reported numerically for all trained deep configurations in \Cref{tab:deep_results}.
The strongest model is DINOv3 Large with cross view gated attention, E2, at $0.8338\pm0.0503$.
DINOv3 Large with two gated depthwise blocks, B5, follows at $0.8274\pm0.0443$.
Under this same fusion head, DINOv2 Large, B4, reaches $0.7902\pm0.0358$.
The matched head comparison adds 0.0372 $R^2$.
It is the closest available backbone comparison because the task head and evaluation protocol are fixed.
The patch size, input resolution, backbone generation and pretraining data still change, so the gain is not an isolated causal effect of one design choice.

The representative no metadata sequence in \Cref{fig:overview}a rises from EfficientNet B3 at 0.6949 to VMamba Base at 0.7432, DINOv2 Large at 0.7902 and DINOv3 Large at 0.8274.
This is an observed backbone progression, not an isolated scaling law, because architecture, resolution, fusion and pretraining change across the sequence.
Smaller frozen probes are weaker.
DINOv2 Base and Giant supervised probes reach 0.6967 and 0.7149.
EfficientNet B3 reaches 0.6949 under the corrected dual view protocol.
VMamba Base gives 0.7432 without metadata and 0.7382 with metadata.
These results show that parameter count alone is not a reliable ordering for this task.

\begin{table}[!t]
\centering
\caption{Deep model results on the locked five fold split. $R^2_w$ is the mean raw scale score across outer folds. Fold SD is the population standard deviation across those folds. Best mean in \textbf{bold}.}
\label{tab:deep_results}
\resizebox{\columnwidth}{!}{%
\begin{tabular}{cllccc}
\toprule
\textbf{ID} & \textbf{Backbone} & \textbf{Fusion or readout} & \textbf{Meta} & \textbf{$R^2_w\uparrow$} & \textbf{Fold SD} \\
\midrule
B2  & EfficientNet B3 & 2$\times$ GDWC & 0  & 0.6949 & 0.0404 \\
B3  & DINOv2 Base & Supervised probe & 0  & 0.6967 & 0.0371 \\
B4  & DINOv2 Large & 2$\times$ GDWC & 0  & 0.7902 & 0.0358 \\
B5  & DINOv3 Large & 2$\times$ GDWC & 0  & 0.8274 & 0.0443 \\
B6  & VMamba Base & Cross scan Mamba & 23 & 0.7382 & 0.0408 \\
B7  & DINOv3 Large & 2$\times$ CVGA & 23 & 0.8277 & 0.0496 \\
B8  & DINOv3 Large & 2$\times$ BidirMamba & 23 & 0.8280 & 0.0414 \\
B9  & VMamba Base & Cross scan Mamba & 0  & 0.7432 & 0.0481 \\
B10 & DINOv2 Giant & Supervised probe & 0  & 0.7149 & 0.0600 \\
E1  & DINOv3 Large & 2$\times$ BidirMamba & 0  & 0.8187 & 0.0500 \\
\textbf{E2} & \textbf{DINOv3 Large} & \textbf{2$\times$ CVGA} & \textbf{0} & \textbf{0.8338} & \textbf{0.0503} \\
E3  & DINOv3 Large & 2$\times$ GDWC & 23 & 0.8286 & 0.0457 \\
E4  & DINOv3 Large & Identity addition & 0  & 0.8181 & 0.0540 \\
E5  & DINOv3 Large & 2$\times$ FullMamba & 0  & 0.7930 & 0.0342 \\
E6  & DINOv3 Large & 1$\times$ GDWC & 0  & 0.8213 & 0.0346 \\
E7  & DINOv3 Large & 4$\times$ GDWC & 0  & 0.8136 & 0.0392 \\
E8  & DINOv3 Large & Identity addition & 23 & 0.8273 & 0.0532 \\
\bottomrule
\end{tabular}%
}
\end{table}

\Cref{fig:overview}b gives a stronger lower bound than a deep model comparison alone and \Cref{tab:classical_results} lists the corresponding values.
Metadata XGBoost, C2, reaches 0.7199 and metadata Ridge, C1, reaches 0.6827.
Handcrafted image features are much weaker at 0.3503 with Ridge and 0.4242 with XGBoost.
Adding metadata raises the handcrafted XGBoost model by 0.1908 to 0.6150 but it remains 0.1048 below metadata XGBoost.
Frozen DINOv3 Large features with Ridge reach 0.7400, an increase of 0.0627 over frozen DINOv2 Base.
PCA followed by XGBoost reduces the DINOv3 result by 0.0850.
The 0.0938 gap between C7 and E2 measures the combined value of image model adaptation, spatial token processing and learned cross view interaction.

\begin{table}[!t]
\centering
\caption{Classical and frozen feature baseline results on the locked five fold split. $R^2_w$ and Fold SD use the same outer folds as \Cref{tab:deep_results}. Best mean in \textbf{bold}.}
\label{tab:classical_results}
\resizebox{\columnwidth}{!}{%
\begin{tabular}{cllcc}
\toprule
\textbf{ID} & \textbf{Input} & \textbf{Learner} & \textbf{$R^2_w\uparrow$} & \textbf{Fold SD} \\
\midrule
C1 & Metadata & Ridge & 0.6827 & 0.0312 \\
C2 & Metadata & XGBoost & 0.7199 & 0.0289 \\
C3 & Handcrafted image & Ridge & 0.3503 & 0.0817 \\
C4 & Handcrafted image & XGBoost & 0.4242 & 0.0814 \\
C5 & Handcrafted image and metadata & XGBoost & 0.6150 & 0.0527 \\
C6 & Frozen DINOv2 Base & Ridge & 0.6773 & 0.0252 \\
\textbf{C7} & \textbf{Frozen DINOv3 Large} & \textbf{Ridge} & \textbf{0.7400} & \textbf{0.0203} \\
C8 & Frozen DINOv3 Large, PCA32 & XGBoost & 0.6550 & 0.0601 \\
\bottomrule
\end{tabular}%
}
\end{table}

\subsection{Fusion and Metadata}

\begin{figure}[!t]
\centering
\includegraphics[width=\columnwidth]{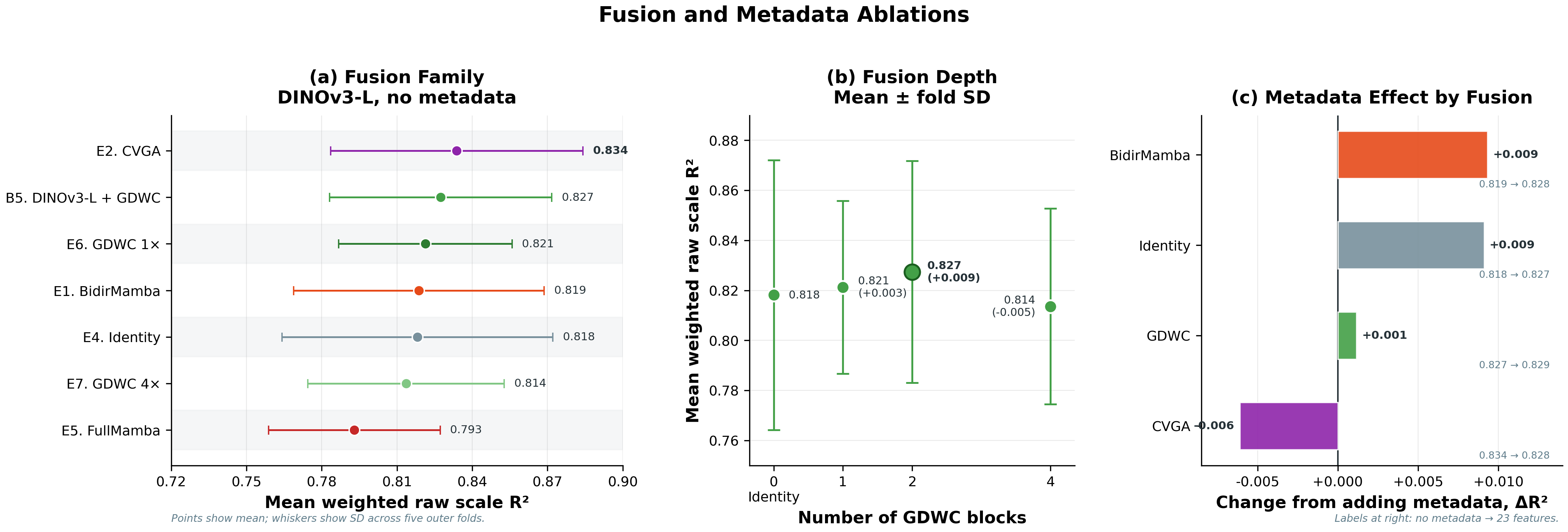}
\caption{Cross view fusion and metadata comparisons. Panel (a) compares fusion families. Panel (b) compares identity and one, two, or four gated depthwise blocks. Panel (c) shows the score change after adding 23 metadata features to four matched image pathways.}
\label{fig:fusion}
\end{figure}

Fusion performance is non-monotonic in architectural complexity, as shown in \Cref{fig:fusion}a.
Full Mamba reaches 0.7930.
Four gated depthwise blocks reach 0.8136, while identity addition reaches 0.8181.
One gated depthwise block gives 0.8213 and two blocks give 0.8274.
Cross view gated attention is best at 0.8338.
The difference between E2 and B5 is 0.0064, which is much smaller than their fold variation.

\Cref{fig:fusion}c visualizes the metadata effects for four matched pairs and \Cref{tab:metadata_pairs} gives their exact values.
Cross view gated attention changes from 0.8338 without metadata to 0.8277 with metadata, a decrease of 0.0061.
Bidirectional Mamba changes from 0.8187 to 0.8280, an increase of 0.0093.
Identity addition changes from 0.8181 to 0.8273, an increase of 0.0091.
Gated depthwise convolution with metadata reaches 0.8286, only 0.0012 above the no metadata counterpart.
The classical baselines show that metadata is predictive but the deep model does not always convert that signal into a better score.
Before metadata, the four matched pathways have a mean of 0.8245 and a range of 0.0157.
With metadata, they have a mean of 0.8279 and a range of 0.0013.
Metadata reduces the between pathway range by 91.7 percent while changing the mean by only +0.0034.
This is a metadata equalization pattern on this split.
The interaction still depends on the image representation and fusion path.

\begin{table}[!t]
\centering
\caption{Matched metadata comparison for DINOv3 Large. $\Delta$ is the mean fold $R^2_w$ with metadata minus the score without metadata.}
\label{tab:metadata_pairs}
\resizebox{\columnwidth}{!}{%
\begin{tabular}{lccc}
\toprule
\textbf{Fusion} & \textbf{No metadata} & \textbf{With metadata} & \textbf{$\Delta$} \\
\midrule
2$\times$ CVGA & 0.8338 (E2) & 0.8277 (B7) & $-0.0061$ \\
2$\times$ BidirMamba & 0.8187 (E1) & 0.8280 (B8) & $+0.0093$ \\
Identity addition & 0.8181 (E4) & 0.8273 (E8) & $+0.0091$ \\
2$\times$ GDWC & 0.8274 (B5) & 0.8286 (E3) & $+0.0012$ \\
\bottomrule
\end{tabular}%
}
\end{table}

\subsection{Fold and Target Stability}

\begin{figure}[!t]
\centering
\includegraphics[width=\columnwidth]{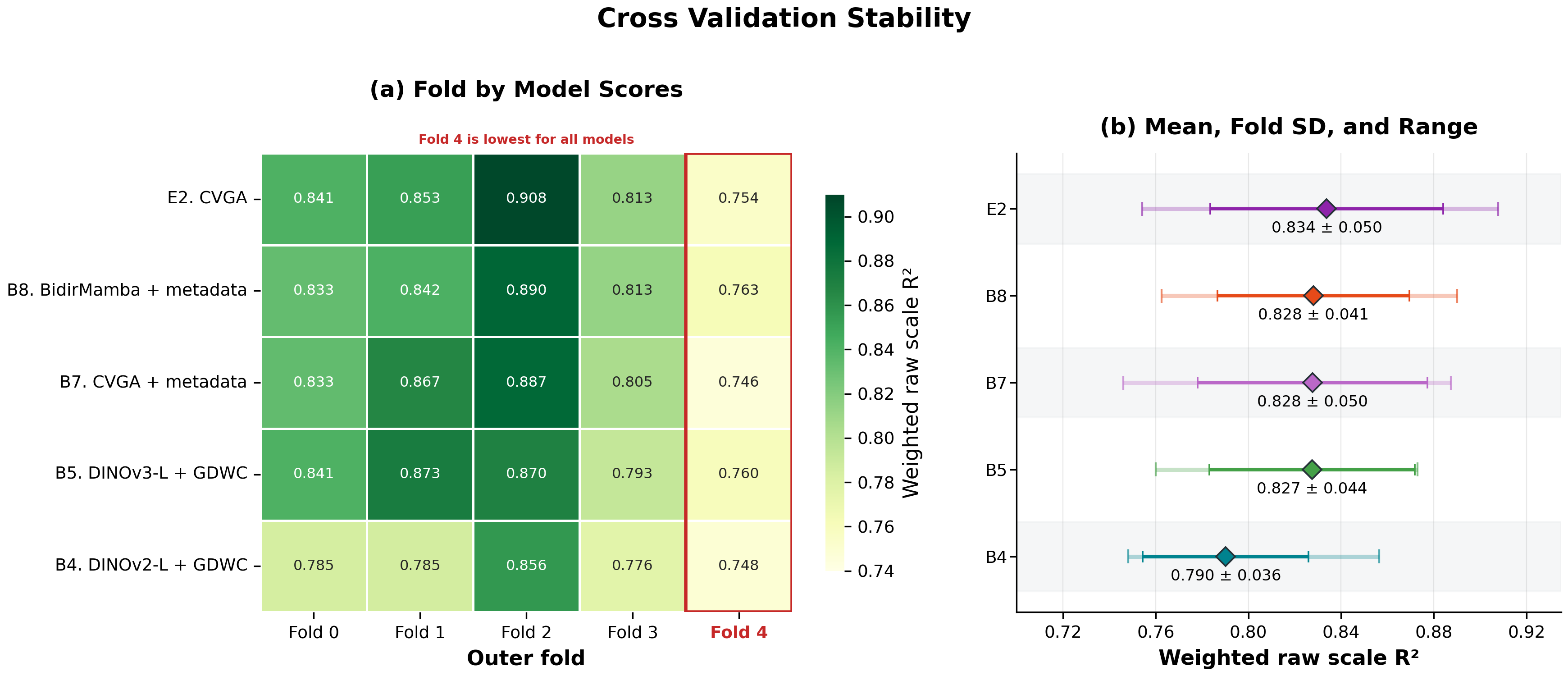}
\caption{Outer fold stability. Panel (a) shows fold scores for selected models. Panel (b) shows each model mean, outer fold standard deviation and range.}
\label{fig:folds}
\end{figure}

Fold results in \Cref{fig:folds} show a common pattern.
Fold 2 is strongest for E2 at 0.9077, while fold 4 is weakest at 0.7541.
Fold 4 is also the lowest fold for B5, B4 and C7.
This shared decline suggests that the partition contains a harder combination of biomass composition or acquisition conditions.
It does not by itself identify the cause because the sample count is small and the fold construction balances only Dry Total strata and image groups.

\begin{figure}[!t]
\centering
\includegraphics[width=\columnwidth]{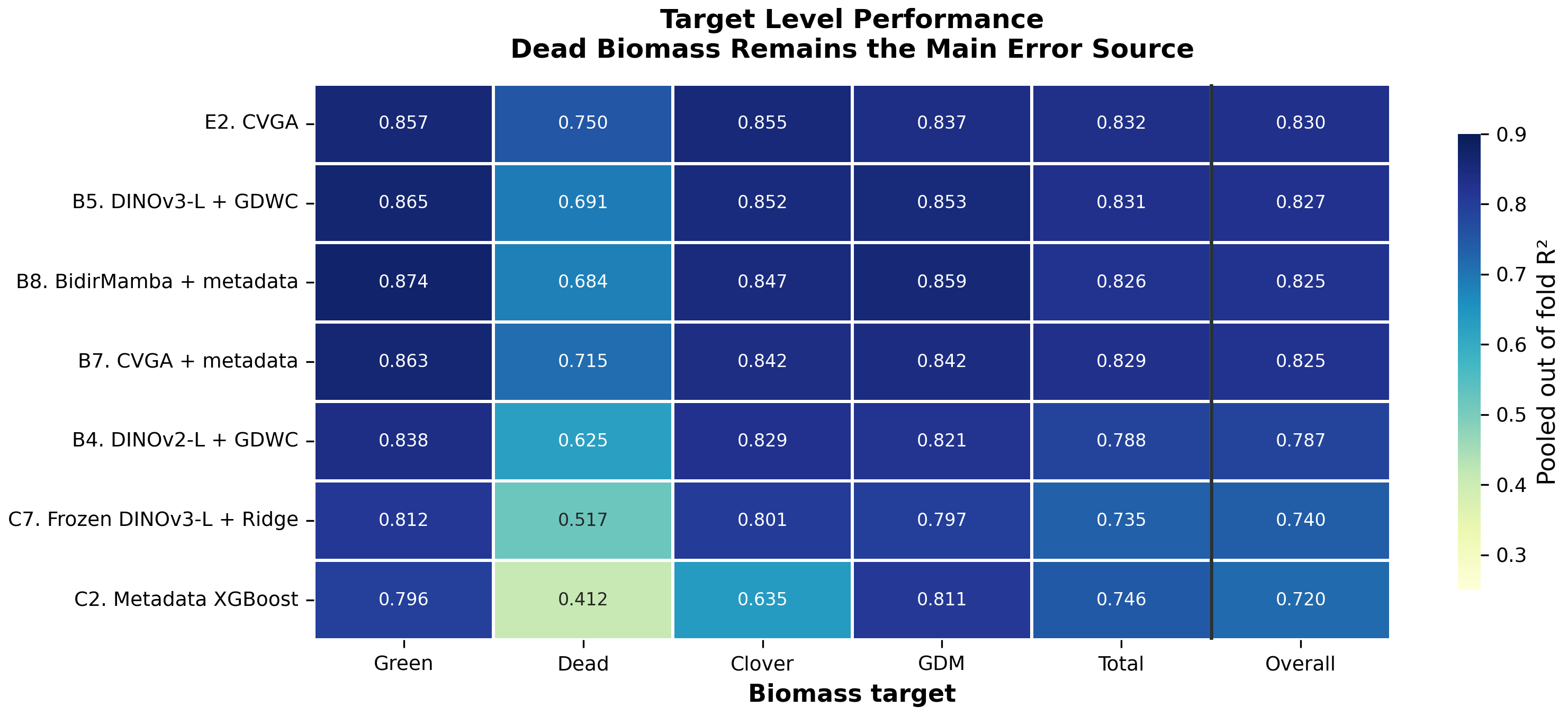}
\caption{Pooled out of fold $R^2$ by biomass target for selected deep, classical and frozen feature models. Overall values use the weighted raw scale score in \Cref{eq:metric}.}
\label{fig:targets}
\end{figure}

Pooled target scores in \Cref{fig:targets} show that Dry Dead is the hardest component.
E2 reaches 0.7503 for Dry Dead, compared with 0.8568 for Dry Green, 0.8545 for Dry Clover, 0.8374 for Green Dry Matter and 0.8317 for Dry Total.
The same difficulty appears in classical and frozen models.
For example, metadata Ridge reaches 0.2786 for Dry Dead and frozen DINOv3 Ridge reaches 0.5170.
This target likely requires more than colour because senescent material is occluded, fragmented and visually mixed with soil and green vegetation.

\subsection{Spatial Attribution}

\begin{figure}[!t]
\centering
\includegraphics[width=\columnwidth]{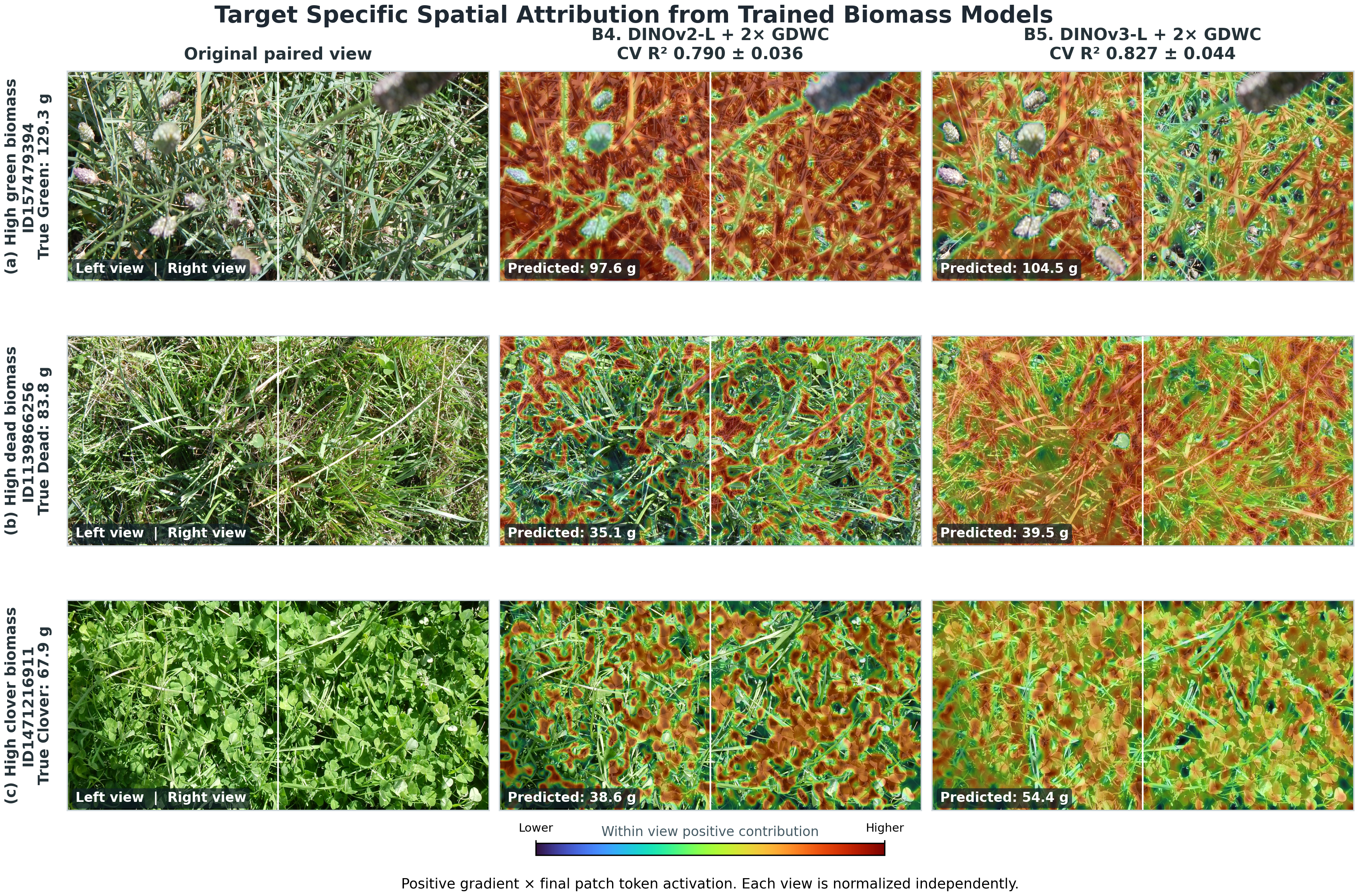}
\caption{Positive gradient times final patch token activation for matched DINOv2 Large and DINOv3 Large models. Each map is normalized independently.}
\label{fig:attribution}
\end{figure}

The final patch token activation is multiplied by its positive input gradient to form a spatial sensitivity map.
Gradient based localization methods use target gradients to identify influential spatial features~\cite{selvaraju2017gradcam}.
Transformer relevance work also shows that attention weights alone do not establish feature relevance~\cite{chefer2021transformer}.
\Cref{fig:attribution} compares DINOv2 Large and DINOv3 Large models with the same gated depthwise fusion head on representative images.
Both models respond to vegetation regions instead of one fixed image position.
DINOv3 often gives more concentrated responses around dense plant structure, while DINOv2 responses are broader in the shown cases.
The maps are qualitative.
They indicate local sensitivity, not causal importance and they do not establish that one backbone reasons more correctly.

\section{Discussion and Limitations}
\label{sec:discussion}

The strongest result comes from combining a large self supervised backbone with direct bidirectional interaction between the two views.
The gain over two gated depthwise blocks is small but the attention model is consistently competitive across folds.
The larger and clearer result is the matched DINOv2 to DINOv3 gain.
This suggests that representation quality is a more stable design axis than adding depth to the fusion module.
The broader monotonic backbone sequence is descriptive instead of a scaling law because several model and training factors change together.

The fusion comparison also corrects a tempting but unsupported explanation.
The results do not show a general ordering between fusion complexity and performance.
Identity, local convolution, attention and state space modules do not lie on one meaningful complexity scale and the best attention model is more global than the convolutional alternatives.
The conclusion is that more elaborate sequence processing does not guarantee better regression on the 357 image CSIRO dataset.

Metadata should also be interpreted carefully.
Its strong classical performance proves that it contains target information when available.
Its mixed effect in deep models may arise from redundant information, optimization imbalance, or fold specific correlations.
Metadata enabled models also assume that the same 23 variables exist at evaluation time, unlike the public challenge setting.
Because metadata is present for every held out sample, these results do not test missing metadata at evaluation time.
The supported finding is score equalization across the four tested pathways, not failure caused by an absent evaluation modality.

\subsection{Result Provenance}

All deep models use the same raw scale weighted $R^2_w$ definition on the locked five fold split.
B6, B7, B8, E3 and E8 use all 23 metadata features, while the remaining deep models use images only.
Among the classical baselines, C1 and C2 use metadata alone and C5 combines metadata with handcrafted image features.
The classical values use nested cross validation on the same outer folds.

\textbf{Limitations:} This study uses one dataset and one locked split.
All reported deep models use one master seed, 17, so small differences lack training seed uncertainty estimates.
We reuse outer folds for checkpoint selection, which can make cross validation estimates optimistic.
The backbone comparison changes model scale, input resolution and pretraining data together.
It measures complete model choice instead of isolating the effect of one backbone change.
Validation on emerging agricultural benchmarks with similar ground truth quality is left to future work.

\section{Conclusion}
\label{sec:conclusion}

This paper evaluates backbones, cross view fusion and metadata for pasture biomass regression under one common protocol.
DINOv3 Large with cross view gated attention is the strongest verified single model at $0.8338\pm0.0503$ mean fold raw $R^2$.
A matched DINOv3 backbone improves substantially over DINOv2, while fusion depth and module complexity do not give a monotonic benefit.
Metadata is valuable for classical prediction and equalizes four matched deep pathways but its effect on each pathway changes direction.
Frozen foundation features provide strong linear baselines but adapted foundation models with token level cross view interaction perform better.
The practical lesson is to measure backbone, fusion and metadata as separate factors, use strong classical controls and report uncertainty at the level that was evaluated.

{\raggedright\noindent Code: \url{https://github.com/WhiteMetagross/FusionComplexityPastureBiomass}\par}

\section*{Acknowledgment}

The author thanks the CSIRO Pasture Biomass dataset contributors and the open source research community whose models and software supported this work.

\IEEEtriggeratref{33}

\end{document}